\documentclass[
]{ceurart}

\usepackage{listings}
\usepackage{cleveref}
\usepackage{multirow}
\usepackage{makecell}
\usepackage{subcaption}
\usepackage{acro}
\DeclareAcronym{iou}{
  short = IoU,
  long  = Intersection over Union
}

\DeclareAcronym{ros}{
  short = ROS,
  long  = Robot Operating System
}
\DeclareAcronym{rsass}{
  short = RSASS,
  long = Robotics Software Architecture-based Self-adaptive Systems
}

\DeclareAcronym{uav}{
    short = UAV,
    long = Unmanned Aerial Vehicle
}
\newcommand{\ourartifact}{SUNSET}
\newcommand{\revtext}[1]{\textcolor{black}{#1}}
\begin{document}

\copyrightyear{2026}
\copyrightclause{Copyright for this paper by its authors.
  Use permitted under Creative Commons License Attribution 4.0
  International (CC BY 4.0).}

\conference{RoSE'26: International Workshop on Robotics Software Engineering}

\title{SUNSET - A Sensor-fUsioN based semantic SegmEnTation exemplar for ROS-based self-adaptation}

\author[1]{Andreas Wiedholz}[%
orcid=0000-0003-0118-8765,
email=andreas.wiedholz@xitaso.com,
]
\cormark[1]
\address[1]{XITASO GmbH, Augsburg, Germany}

\author[2]{Rafael Paintner}[%
]
\address[2]{German Aerospace Center (DLR) Institute of Flight Systems}

\author[1]{Alwin Hoffmann}[]
\author[3]{Carlos Hernandez}[]
\address[3]{TU Delft, Department of Mechanical Engineering}
\author[1]{Tobias Huber}[]

\cortext[1]{Corresponding author.}

\begin{abstract}
The fact that robots are getting deployed more often in dynamic environments, together with the increasing complexity of their software systems, raises the need for self-adaptive approaches.
In these environments robotic software systems increasingly encounter (1) failures whose symptoms are easy to observe but root causes might be ambiguous or (2) multiple failures appearing concurrently. 
We present \ourartifact{}, a ROS2-based exemplar that enables rigorous, repeatable evaluation of architecture-based self-adaptation in such conditions. 
It implements a sensor fusion semantic-segmentation pipeline driven by a trained Machine Learning (ML) model whose input preprocessing can be perturbed to induce realistic performance degradations.
The exemplar exposes five observable failures, each of which can be caused by different faults and supports concurrent failures spanning self-healing and self-optimisation.
\ourartifact{} includes the segmentation pipeline, a trained ML model, fault-injection scripts, a baseline controller for further comparisons, and step-by-step integration and evaluation documentation to facilitate reproducible studies. The code is available at \url{https://github.com/XITASO/sunset}.
\end{abstract}


\maketitle

\section{Introduction}
Robotic systems increasingly operate in dynamic, unmapped environments (e.g., \ac{uav}), which raises the need for adaptation.
The research area of \ac{rsass} aims to deal with the challenges accompanying this development by adapting the software during runtime \cite{alberts_software_2025}.


%
It is often much easier to detect failures of a system than their causes, which might be be an internal fault or an external perturbation (both of which we will refer to as "fault" in the remaining paper for simplicity). 
However, since the true root cause of the failure is often unknown, this raises uncertainty in the overall system.
We focus on two challenging cases that are particularly relevant for robotic systems:
(1) symptoms of failures that do not map cleanly to a single root cause, and (2) concurrent failures that appear together \,---\, either raised independently by multiple faults or as ripple effects.
\begin{figure}
    \centering
    \includegraphics[width=\linewidth]{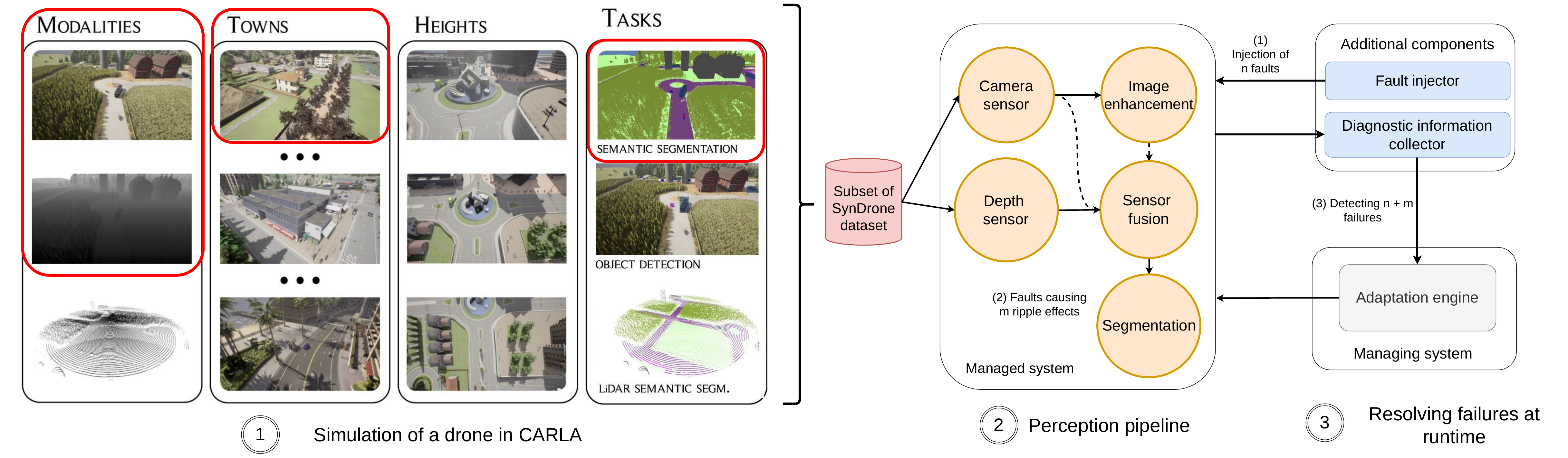}
    \caption{\ourartifact{} - A sensor fusion based semantic segmentation pipeline based on the SynDrone dataset\cite{syndrone_dataset} with a fault injector for testing self-adaptive ROS-based systems. During experiments multiple faults will be injected which may cause ripple effects. The managing system detects multiple failures at the same time and has to adapt the managed system accordingly.}
    \label{fig:teaser}
\end{figure}
While general self-adaptive systems have studied concurrent failures\cite{vogel_mrubis_2018}, \ac{rsass} lacks exemplars that evaluate self-adaptivity under concurrent, cause-ambiguous failures.

To fill this gap, this paper introduces \ourartifact{} \,---\, an exemplar which implements a Sensor fUsioN based Semantic sEgmentaTion pipeline (see \Cref{fig:teaser}).  
To make \ourartifact{} suitable for evaluating self-adaptive approaches in robotics, we chose a common and realistic robotic task (semantic segmentation) \cite{tzelepi_semantic_2021} and implement it in the \ac{ros} the main technology used for \ac{rsass}\cite{alberts_software_2025}. 
\ourartifact{} supports external self-adaptation with a clearly separated managed and managing system \cite{kephart_vision_2003}. 
This separation simplifies the integration of different managing systems on the same managed system and follows common practice in robotic and \ac{ros}-based self-adaptivity exemplars \cite{gil_body_2021, silva_suave_2023, imrie_aloft_2024}.

For the semantic segmentation, \ourartifact{} implements a sensor-fusion-based approach that uses a real machine-learning model whose inputs can be perturbed to produce realistic degradations in performance. 
Moreover, the exemplar supports the four main types of architectural adaptations common in \ac{rsass}\cite{alberts_software_2025} which are aimed at either self-optimization or self-healing, i.e., more critical failures, capabilities.

This combination of realistic \ac{ros}{2}-based robotics software integration, rich adaptation capabilities, and concurrent failures with unknown sources makes \ourartifact{} a novel and challenging exemplar for evaluating \ac{rsass} approaches.
To summarize, our work offers the following contributions:
\begin{itemize}
    \item An exemplar in which multiple faults can lead to the same failure requiring root cause analysis. 
    \item The first \ac{ros}-based exemplar that allows all common adaptations for self-adaptive robotic applications to be performed by a managing system to resolve concurrent failures.
    \item A ML model offering realistic data for degraded performance.
    \item An exemplar with detailed documentation how to evaluate new managing systems with an exemplary baseline.
\end{itemize}

\section{Related work}
\ac{rsass} research has been present for at least 15 years with \ac{ros} being the main common ground in this field.
Many of these approaches are focused on their respective use case, causing many evaluation scenarios to be tailored to the proposed managing system itself \cite{alberts_software_2025}.
This makes it hard to prove that one method can be implemented for multiple use cases.
Therefore, more exemplars allowing \ac{ros}-based self-adaptive approaches to be evaluated on are required.

A detailed overview of self-adaptive robotic exemplars is given in Robomax \cite{askarpour_robomax_2021} which contains 12 exemplars of different robotic applications and details of the robotic mission and the environment.
Apart from the Body Sensor Network \cite{gil_body_2021} \,---\, a patient health status monitoring application implemented in \ac{ros}1 \,---\, the exemplars focus on theoretical scenarios instead of empirical exemplars.


\citet{silva_suave_2023} introduce SUAVE which simulates a \ac{ros}2-based underwater robot with the mission to detect and inspect a pipeline.
In the extended version, SUAVE contains 3 failures, namely environmental factors, the need to recharge a battery or the breakdown of one of the robot's thrusters.
All of the failures in SUAVE occur sequentially and have a clearly defined adaptation \revtext{spanning self-healing and self-optimization} concerns that resolves it.
\revtext{The main difference to \ourartifact{} is that in SUAVE the adaptation to resolve a detected failure is always the same.}

An exemplar for drone control implemented in \ac{ros}1 has been implemented by \citet{imrie_aloft_2024}.
The use of Gazebo allows a physically accurate simulation and contains physical concerns as well as time-related concerns raising the need for self-adaptation.
This can be related to failing components in the system or environmental factors necessitating changes in the plans.

\revtext{To the best of our knowledge}, \ourartifact{} is the first \ac{ros}-based exemplar that supports concurrent failures with unknown causes necessitating a managing system to prioritise between detected failures.
It is implemented in \ac{ros}2 instead of the majority of the \ac{ros}1-based related work \cite{askarpour_robomax_2021, imrie_aloft_2024, gil_body_2021} and enables managing systems to use all adaptations that are common in \revtext{\ac{rsass}} \cite{alberts_software_2025}.
This enables robotics research to evaluate managing systems with enhanced capabilities on an independent exemplar.

\section{\ourartifact}

\subsection{Use-Case Description}
\ourartifact{} models the perception of an \ac{uav} that must semantically understand its surroundings during a mission \,---\, a core capability in many modern robotic systems \cite{tzelepi_semantic_2021}. 
As data basis for our exemplar, we utilise the SynDrone Dataset (Town 01)\cite{syndrone_dataset}, which simulates a \ac{uav} in a CARLA simulation containing both urban and rural scenes (see \cref{fig:teaser}).
The \ac{uav} is equipped with RGB and depth cameras, and the resulting data is continuously fused and interpreted by a semantic segmentation model to support tasks such as safe navigation and environment mapping.

\subsection{System Architecture}
The \ac{ros} architecture of the semantic segmentation pipeline in \ourartifact{} consists of standard components for sensor fusion (see \Cref{fig:teaser}).
A \textbf{camera sensor} and \textbf{depth sensor} node read RGB and depth images from a \ac{ros} bag that contains pre-recorded flight data and provide them to the rest of the pipeline.
Using pre-recorded data rather than live simulations improves the replicability of \ourartifact{} experiments, as the same sensor inputs can be replayed across different runs and configurations.
\ourartifact{} also entails an \textbf{image enhancement} node that is able to reverse a degradation of RGB images.
The two data modalities are aligned by a \textbf{sensor fusion} node that matches RGB–D pairs based on the timestamps of the received images.
The fused (or single-modality) images are then processed by a \textbf{segmentation} node that performs semantic segmentation.
For each configuration (RGB, Depth, RGB+Depth), we train a separate ML model from scratch on the SynDrone dataset\cite{syndrone_dataset}, while keeping the rest of the mission scenario unchanged.
For the model architecture, we use the ResNet50 based Deeplab-V3 model with the late fusion strategy tested by \citet{syndrone_dataset}.

Alongside the segmentation pipeline, a \textbf{fault injector} introduces faults (see \Cref{ssec::uncertainties}) directly into the \ac{ros} nodes.
Additionally, we provide a \ac{ros} node that collects information needed by the managing system to detect failures and therefore the necessity of an adaptation of the managed system, e.g., degraded performance of the segmentation model.
For the managing system we propose a baseline (see \Cref{ssec::baseline}), which can be substituted with any other \ac{ros}{2}-based managing system.

\subsection{Faults}
\label{ssec::uncertainties}
\begin{figure*}
\centering
    \begin{subfigure}[t]{0.5\textwidth}
        \centering
        \includegraphics[width=\textwidth]{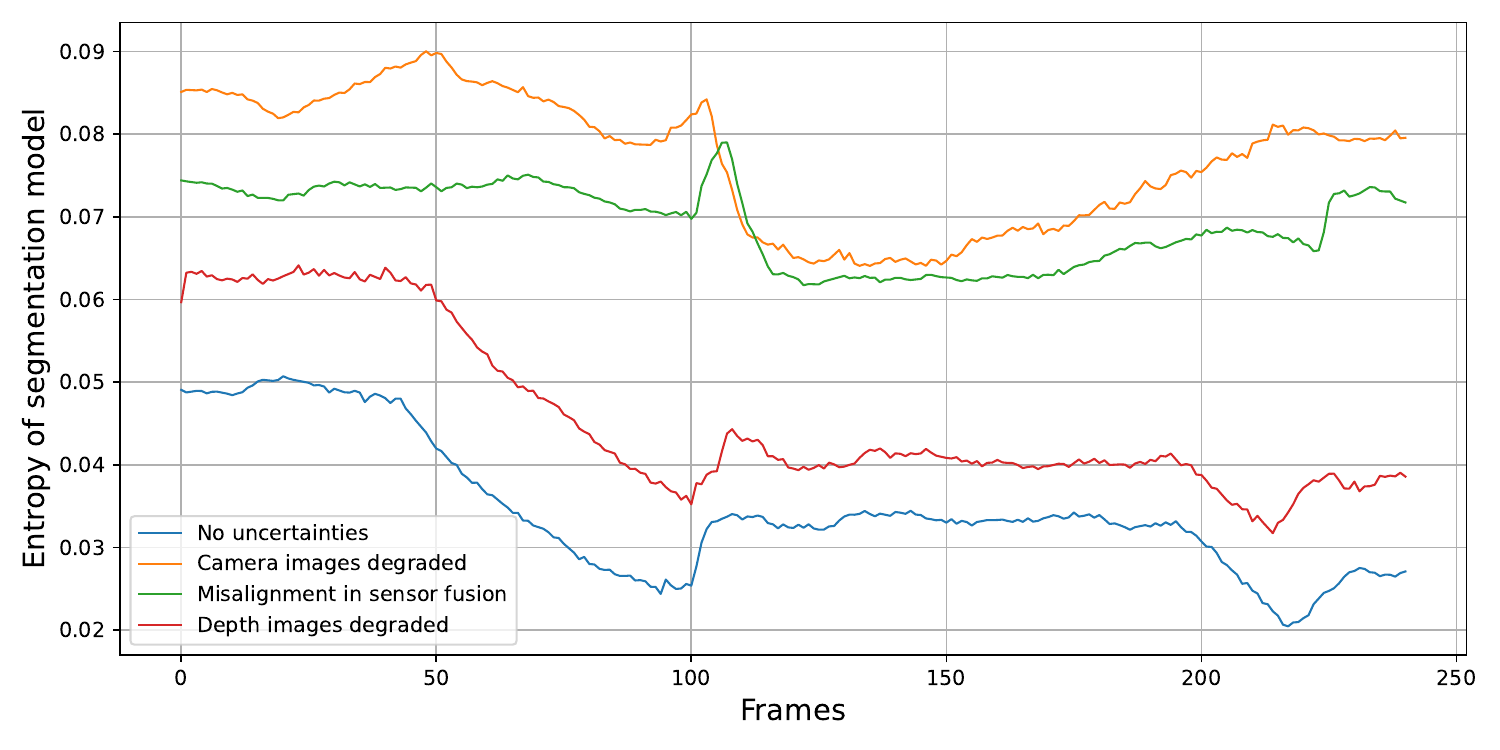}
        \caption{The effect of different faults on the entropy of our segmentation model. This refers to failure F4.}
        \label{fig:comparison-uncertainties}
    \end{subfigure}%
    ~ 
    \begin{subfigure}[t]{0.5\textwidth}
        \centering
        \includegraphics[width=\textwidth]{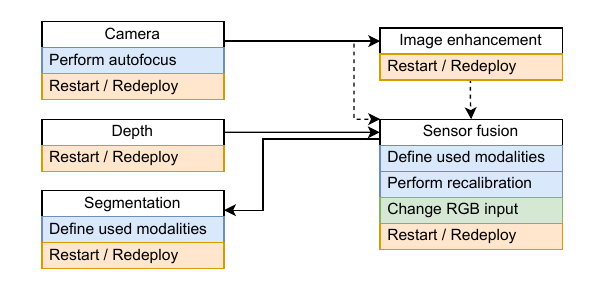}
        \caption{Potential adaptations in \ourartifact{}. Blue := Reparametrization, Green := Change of communication, Orange := Software/Hardware restart}
        \label{fig:adaptations_overview}
    \end{subfigure}
    \caption{Overview of the characteristics of \ourartifact{}}
\end{figure*}
Our fault injector can introduce 11 faults with every fault being the reason for a failure.
The faults in \ourartifact{} result in 5 different failures (\textit{F1-5}) that require self-healing (\textit{F1-3}) and self-optimising (\textit{F4-5}) capabilities.
Each failure can be solved with at least one of the possible adaptations.
Resolving all injected faults requires all of the adaptations our system is capable of (see \Cref{ssec:adaptations}).
Categorised by their failure, the introduced faults are:


\textbf{F1-3: Communication channel outage} These failures can be produced by the RGB camera (\textit{F1}), sensor fusion (\textit{F2}), or the segmentation node (\textit{F3}) and can be detected by monitoring the frequency of their respective topic. 
For these failures, \ourartifact{} simulates 6 faults: hardware disconnect of the camera, issues in the driver, memory leakage or synchronization issue in the sensor fusion or GPU failure and CUDA out of memory exception in the segmentation node.
All of these faults require different adaptations \,---\, lifecycle restart or node redeploy \,---\, but result in the same kind of failure in the system.

\textbf{F4: Reduced segmentation performance} 
Since we use a trained ML model for segmentation, we are able to measure realistic effects on the uncertainty of the model in its predictions during runtime when degrading the input.
We achieve this by calculating the mean entropy of the models' segmentation logits.
If the entropy increases, i.e., the model gets uncertain about its predictions, the quality of the segmentation decreases. 
As the segmentation is dependent on multiple nodes, \ourartifact{} simulates four faults that can lead to this symptom (see \Cref{fig:comparison-uncertainties}).
First, a degraded image quality is simulated by shifting the colour space of the images provided to the camera node.
Second, if the image enhancement is applied on non-degraded images, it leads to degraded images and has the same effect on the entropy of the segmentation model. 
Third, we simulate misalignment during fusion of the RGB and depth images by shifting one of the images, which results in a spatial misalignment between RGB and depth.
Lastly, the depth images can be degraded with Gaussian noise.

\textbf{F5: Blurred camera image} An incorrect camera focus is simulated by blurring the image.



\subsection{Possible adaptations}
\label{ssec:adaptations}
\begin{figure*}
    \centering

\end{figure*}
In order to resolve failures in \ourartifact{}, i.e., identifying the fault and recovering from it, we can perform the four main adaptations in \ac{rsass}\cite{alberts_software_2025} which are described in the following.
\Cref{fig:adaptations_overview} gives an overview of all the possible adaptations per \ac{ros} node in \ourartifact{} that can be used to resolve runtime failures.
In the following we detail the behaviour of the \ac{ros} nodes in \ourartifact{} during an adaptation.

\textbf{Reparametrisation}
In order to reparametrise a \ac{ros} node, we use the standard \ac{ros} parameters.
Additionally to setting a new value, these parameters can be used in some nodes, e.g., the camera or sensor fusion, to perform an action aimed for self-optimisation when set to True.


\textbf{Change of communication}
The \ac{ros} node can change any of its subscribed topics at runtime in the following way:
After the managing subsystem provides the name of the new topic, the adapted \ac{ros} node destroys its current subscription and creates a new one with the same message type.

\textbf{Addition / removal of components}
Since our common base class used for all \ac{ros} nodes in \ourartifact{} inherits from the Lifecycle node class, all nodes can be within four predefined lifecycle states during runtime. 
We consider a node added to the system if it is in the \textit{ACTIVE} state and removed if it is in any other state.
Depending on the injected fault, \textit{F1-3} can be resolved by restarting the respective node, i.e., removing and adding it to the system.

\textbf{Redeploy}
If a node is redeployed, the \ac{ros} node with the OS process is completely terminated simulating a hardware restart.
To simulate this, an artificial delay is introduced when redeploying the respective nodes, e.g. camera.
In general, redeploying a node typically resolves all \ac{ros} node internal fault; however failures with external causes, e.g., image degradation remain persistent.

\subsection{Expanding \ourartifact{}}
Adding new image domains to SUNSET is straight forward.
For example, for the SynDrone dataset we already provide a script that can add additional video sequences from different towns and heights (see \Cref{fig:teaser}).  
\ourartifact{} can also be extended to other common robotic tasks like object detection, LiDAR-based segmentation or even control with the original \ourartifact{} embedded as perception part of a complete robotic system.
To facilitate the implementation of new nodes, we provide a base class that extends the \ac{ros} lifecycle node class and implements the behaviour of all described adaptations.



\section{Evaluation}

To measure the performance of the managing system and the impact on the managed system, we provide several metrics which we calculate based on log files.
The performance of the managed system can be measured by the quality of the segmentation model's predictions (\ac{iou}) over all frames and the availability of the overall managed system.
For the availability of the managed system, we introduce the system downtime that accumulates the time that the segmentation algorithm didn't send new data in the expected frequency. 
This indicates that one component had an outage.
For context, one experiment runs for 20 seconds, i.e., a downtime of 5 seconds would result in a system availability of 75\%.
Other metrics we provide solely focus on the performance of the managing system:
\begin{itemize}
    \item Ratio of resolved failures and executed adaptations $\frac{f_{resolved}}{a_{executed}}$: How many failures were resolved per executed system adaptation?
    This metric measures how many concurrent failures a single system adaptation solves.
    \item Reaction time $t_\text{react}$: Since for each detected failure, there are multiple faults potentially being responsible, this metric measures the time it took to detect a symptom and select the correct adaptation of the system.
    \item Number of unnecessary redeploys $\text{\#redeploy}_\text{u}$: Number of redeploys that were executed even though a cheaper adaptation would have sufficed.
\end{itemize}
\label{ssec::baseline}
\begin{table*}
\centering
\caption{Baseline performance of SUNSET. Directional arrows specify metric desirability.}
\label{tab:sample-evaluation}
\begin{tabular}{cccccc}
\toprule
\makecell{Managing  system }& $\frac{f_{resolved}}{s_{executed}} \uparrow$ & $t_{react}[s] \downarrow$ & \#$\text{redeploy}_u \downarrow$ & $t_{down}[s] \downarrow$ & IoU $\uparrow$ \\
\midrule
\makecell{None w/o  faults}&      N/A        &    N/A          &  N/A            &   N/A            & 0.47 $\pm$ 0.02 \\
\makecell{None w/ faults (F4 and F5)}&      N/A        &    N/A             &  N/A            &   N/A            & 0.28 $\pm$ 0.08    \\
Baseline  & 0.89 $\pm$ 0.16 & 1.44 $\pm$ 1.32 & 2.02 $\pm$ 1.04 & 6.52 $\pm$ 4.67& 0.30 $\pm$ 0.09 \\
\bottomrule
\end{tabular}
\end{table*}
To provide first results for our metrics, we implement a managing system based on the MAPE-K feedback loop \cite{kephart_vision_2003} with straightforward adaptations to the system.
This can be used by other researchers as baseline to compare their managing system with.
First, the managing system redeploys every node that stops sending data, \revtext{i.e., $f_\text{node} == 0$} (\textit{F1-3}).
Second, it performs a recalibration in the sensor fusion node as soon as the entropy of the segmentation model is higher than 0.06 (\textit{F4}, \revtext{see \Cref{fig:comparison-uncertainties})}.

The results of our baseline are presented in \Cref{tab:sample-evaluation}.
In the first two rows, we present the results for running \ourartifact{} without any faults and with faults present the whole time (except outages, therefore no $t_{down}$) to show the best and worst possible \ac{iou}.
The other metrics are not applicable here since there is no downtime and no managing system present that can be evaluated.
\revtext{In every run one fault for self-healing (F1-3) and two faults for self-optimization (F4-5) are injected simultaneously.
To ensure reproducibility, 54 scenarios were created containing every possible combination of these faults.} 
The last row shows the baseline that ran every scenario three times.
By design the baseline is not able to detect three out of four faults for \textit{F4} which results in a high standard deviation for the \ac{iou}.
As the baseline redeploys every node in which \textit{F1-3} is detected, $\#redeploy_u$ has a high value.

\section{Conclusion \& Future work}
In this work, we present \ourartifact{} an exemplar for evaluating managing systems for self-adaptive robotic systems.
Our exemplar contains a segmentation pipeline with a real ML model and different failures causing degraded performance of the model or entire system outages.
In \ourartifact{}, multiple faults can (1) cause the same failure and (2) occur simultaneously, necessitating a deeper analysis of the system.
Therefore, \ourartifact{} presents a novel and challenging benchmark for evaluating \ac{rsass}.

\begin{acknowledgments}
The authors would like to thank the Federal Ministry of Economic Affairs and Climate Action of Germany for funding the described activities through the LuFo VI-3 program, funding code 20F2201D.
\end{acknowledgments}

\section*{Declaration on Generative AI}
  The author(s) have not employed any Generative AI tools.
  

\bibliography{references}

\end{document}